\pdfoutput=1
\documentclass[A4paper, 11 pt, conference]{ieeeconf}  
\IEEEoverridecommandlockouts

\usepackage{geometry}
\usepackage{graphics} 
\usepackage{epsfig} 
\usepackage{mathptmx} 
\usepackage{times} 
\usepackage{amsmath} 
\usepackage{amssymb}  
\usepackage{subfigure}
\usepackage{algorithm} 
\usepackage{algorithmic} 
\usepackage{multirow} 

\usepackage{subfiles}
\usepackage{xcolor}
\usepackage{caption}  
\usepackage{tabularx} 
\usepackage{booktabs} 
\usepackage{upgreek} 
\usepackage[english]{babel}
\usepackage{hyperref} 

\usepackage{amsmath,amsfonts,bm}

\def\eqref#1{equation~\ref{#1}}

\def\1{\bm{1}}

\def\va{{\bm{a}}}

\def\vs{{\bm{s}}}

\DeclareMathAlphabet{\mathsfit}{\encodingdefault}{\sfdefault}{m}{sl}
\SetMathAlphabet{\mathsfit}{bold}{\encodingdefault}{\sfdefault}{bx}{n}

\def\sA{{\mathbb{A}}}

\def\sS{{\mathbb{S}}}

\usepackage{adjustbox}

\begin{document}

\title{\LARGE \bf
A Survey on Deep-Learning Approaches for Vehicle Trajectory Prediction in Autonomous Driving}

\author{Jianbang Liu$^{*1}$, Xinyu Mao$^{*1}$, Yuqi Fang$^2$, Delong Zhu$^1$, and Max Q.-H. Meng$^{\dagger3}$, \textit{Fellow, IEEE} 
\thanks{$*$Equal contribution. $\dagger$The corresponding author.}
\thanks{$^1$Jianbang Liu, Xinyu Mao and Delong Zhu are with the Department of Electronic Engineering, The Chinese University of Hong Kong, Hong Kong. (\tt\small \{henryliu, maoxinyu, zhudelong\}@link.cuhk.edu.hk)}
\thanks{$^2$Yuqi Fang is with the Department of Biomedical Engineering, The Chinese University of Hong Kong, Hong Kong. (\tt\small fangyuqi@link.cuhk.edu.hk) }
\thanks{
   $^3$Max Q.-H. Meng is with the Department of Electronic and Electrical Engineering of the Southern University of Science and Technology in Shenzhen, China, on leave from the Department of Electronic Engineering, The Chinese University of Hong Kong, Hong Kong, and also with the Shenzhen Research Institute of the Chinese University of Hong Kong in Shenzhen, China. ({\tt\small max.meng@ieee.org}.)
}} 

\maketitle 
\thispagestyle{empty}

\begin{abstract}
With the rapid development of machine learning, autonomous driving has become a hot issue, making urgent demands for more intelligent perception and planning systems. 
Self-driving cars can avoid traffic crashes with precisely predicted future trajectories of surrounding vehicles. 
In this work, we review and categorize existing learning-based trajectory forecasting methods from perspectives of representation, modeling, and learning. 
Moreover, we make our implementation of Target-driveN Trajectory Prediction publicly available at \url{https://github.com/Henry1iu/TNT-Trajectory-Predition}, demonstrating its outstanding performance whereas its original codes are withheld. 
Enlightenment is expected for researchers seeking to improve trajectory prediction performance based on the achievement we have made.
\end{abstract}

\section{Introduction}
\label{sec:intro}
Human depends increasingly on the autonomous system to be freed from tedious tasks, a typical instance of which is autonomous driving.
In the self-driving scenario, safety is the first concern.
Predicting the future trajectories of participants helps the autonomous system find the most promising local path planning solution and prevent possible collision. 

Pioneers have predicted the motion of dynamic objects with Kalman filter\cite{ess2010object}, linear trajectory avoidance model\cite{pellegrini2011predicting}, and social force model\cite{luber2010people}.
Compared to these traditional modeling techniques based on hand-crafted features, deep learning algorithms that learn features automatically via optimizing loss functions have recently attracted researchers' attention.
In this work, we survey some recent approaches for vehicle trajectory prediction and present some innovative ideas.
We implement the approach presented by Zhao et al. in \cite{zhao2020tnt} since its vectorized representation is innovative and efficient. 
By publishing this survey along with our code, we hope that new breakthroughs can be made in this rapidly expanding research field. 

The two main contributions are as follows:
\begin{enumerate}
    \item Recent deep-learning approaches tackling trajectory prediction problems in driving scenarios are reviewed and discussed. 
    \item We implement the prediction model introduced by Zhao et al. \cite{zhao2020tnt} and release our code to the research community. 
\end{enumerate}

\section{Problem Formulation}
\label{sec:problem}

A self-driving system is assumed to be equipped with detection and tracking modules that observe state $\displaystyle \sS$ accurately for all the involved agents $\displaystyle \sA$. 
Given a scene, the prediction target is denoted as $\displaystyle \va_{tar}$ and the surrounding agents are denoted as $\displaystyle \sA_{nbrs} = \{\va_1, \va_2, ..., \va_m\}$.
The state of agent $\displaystyle \va_i \in \sA$ at frame $t$ is denoted as $\displaystyle s^{t}_i$, including features such as position, velocity, heading angle, actor type, and $\displaystyle \vs_i = \{s^{-T_{obs} + 1}_i, s^{-T_{obs} + 2}_i, ..., s^{0}_i\}$ denote the sequence of states sampled at different timestamps throughout the observation period $\displaystyle T_{obs}$.
The objective of a predictive framework is to predict future trajectories $\uptau_{tar} = \{\tau_i | i = 1,2,..., K\}$ of the target agent $\displaystyle \va_{tar}$, where $\displaystyle \tau_i = \{( x^{1}_i,  y^{1}_i), ( x^{2}_i,  y^{2}_i), ..., ( x^{T_{pred}}_i,  y^{T_{pred}}_i)\}$ denotes the predicted trajectories for the target agent up to the prediction horizon $\displaystyle T_{pred}$. 
Besides, the predicted trajectories $\displaystyle \tau_i \in \uptau_{tar}$ need to satisfy the no-collision constraint and target agent's kinematic constraints. 

\section{Methods}
\label{sec:method}
In this section, we review the data representation, model structure, learning techniques, and objective functions of some representative approaches. 

\begin{table*}[h]
	\label{tab:method_summary}
	\centering
	\caption{Comparison of Modality and Modeling: $\displaystyle CSDTS_{multi}$, $\Vec{Polyline}$, $\Delta_{dist}$, $\mathcal{G}_{Lane}$, $\uptau_{cond}$, $Anchor_T$, $\Delta$, $State\_Map^F$, and $\mathcal{L}$ denote CSDTS containing position under multiple coordinate system, the vectorized polyline, the relative distance, the lane connectivity graph,  the conditioned future trajectory, the trajectory anchors, the offset, the future state map, and the Laplacian distribution, respectively. S-Pool indicates the social pooling\cite{alahi2016social, gupta2018social}, Spatial\&Temporal indicates the spatial and temporal learning\cite{ye2021tpcn}, SA and DPA indicate the self-attention and the dot-product attention. XFMR is the abbreviation of transformer.}
    \begin{tabular}{@{}lcccccccc@{}}
    \toprule
		\multirow{2}{*}{}     & \multicolumn{3}{c}{Input Modality}    & Output Modality                                           & \multicolumn{3}{c}{Modeling}                              \\ 
		\cmidrule(lr){2-4} \cmidrule(lr){5-5} \cmidrule(l){6-8} 
		Model & $\displaystyle \sS$ Repre. & Scene Repre. & Extra & Output & $\displaystyle \sS$ Enc. & Scene Enc. & Interaction \\ \midrule
		DESIRE\cite{lee2017desire} & \multicolumn{1}{c|}{CSDTS} & \multicolumn{1}{c|}{BEV} & - & $\uptau_{tar}$, $\Delta$, Score & \multicolumn{1}{c|}{GRU, CVAE} & \multicolumn{1}{c|}{ConvNet} & S-Pool \\
		MTP\cite{cui2019multimodal} & \multicolumn{2}{c|}{BEV} & State & $\uptau_{tar}$, $\Pr(\cdot)$ & \multicolumn{2}{c|}{MobileNetV2\cite{sandler2018mobilenetv2}} & - \\
		MultiPath\cite{chai2019multipath} & \multicolumn{2}{c|}{BEV} & - & $\uptau_{tar}$, $\Pr(\cdot)$, $\mu$, $\sigma$ & \multicolumn{2}{c|}{ResNet\cite{he2016deep}} & CNN \\
		CoverNet\cite{phan2020covernet} & \multicolumn{2}{c|}{BEV} & State & $\uptau_{tar}$, $\Pr(\cdot)$ & \multicolumn{2}{c|}{ResNet\cite{he2016deep}} & - \\
		VectorNet\cite{gao2020vectornet} & \multicolumn{2}{c|}{$\Vec{Polyline}$} & - & $\uptau_{tar}$ & \multicolumn{2}{c|}{PointNet\cite{qi2017pointnet}} & SA-GNN \\
		TPNet\cite{fang2020tpnet} & \multicolumn{2}{c|}{BEV} & - & $P(t)$, $\Delta$, $\Pr(\cdot)$ & \multicolumn{2}{c|}{ResNet\cite{he2016deep}} & - \\
		SAMMP\cite{mercat2020multi} & \multicolumn{1}{c|}{CSDTS} & \multicolumn{1}{c|}{-} & - & $\uptau_{tar}$ & \multicolumn{1}{c|}{LSTM} & \multicolumn{1}{c|}{-} & SA, RNN \\
		Luo\cite{luo2020probabilistic} & \multicolumn{1}{c|}{CSDTS} & \multicolumn{1}{c|}{Lanes, Points} & - & Lanes, $\mu$, $\sigma$ & \multicolumn{1}{c|}{LSTM} & \multicolumn{1}{c|}{1D-Conv, MLP} & DPA \\
		BaiduUS\cite{pan2020lane} & \multicolumn{1}{c|}{CSDTS} & \multicolumn{1}{c|}{Lanes, Points} & $\Delta_{dist}$ & Lanes, $\mu$, $\sigma$ & \multicolumn{1}{c|}{LSTM} & \multicolumn{1}{c|}{LSTM} & ST-G \\
		LaneGCN\cite{liang2020learning} & \multicolumn{1}{c|}{CSDTS} & \multicolumn{1}{c|}{$\Vec{Polyline}$} & $\mathcal{G}_{Lane}$ & $\uptau_{tar}$, Score & \multicolumn{1}{c|}{1D-Conv} & \multicolumn{1}{c|}{LaneGCN} & SA \\
		DATF\cite{park2020diverse} & \multicolumn{1}{c|}{CSDTS} & \multicolumn{1}{c|}{BEV} & - & $\uptau_{tar}$ & \multicolumn{1}{c|}{LSTM} & \multicolumn{1}{c|}{ConvNet} & SA \\ 
		SMART\cite{sriram2020smart} & \multicolumn{2}{c|}{BEV} & - & $State\_Map^{F}$ & \multicolumn{2}{c|}{U-Net, CVAE} & S-Pool \\
		PiP\cite{song2020pip} & \multicolumn{1}{c|}{CSDTS} & \multicolumn{1}{c|}{-} & $\uptau_{cond}$ & $\uptau_{tar}$ & \multicolumn{1}{c|}{LSTM} & \multicolumn{1}{c|}{-} & S-Pool \\
		TNT\cite{zhao2020tnt} & \multicolumn{2}{c|}{$\Vec{Polyline}$} & - & $\uptau_{tar}$, Score & \multicolumn{2}{c|}{PointNet\cite{qi2017pointnet}} & SA-GNN \\
		WIMP\cite{khandelwal2020if} & \multicolumn{1}{c|}{CSDTS} & \multicolumn{1}{c|}{$\Vec{Polyline}$} & - & $\uptau_{tar}$ & \multicolumn{1}{c|}{RNN} & \multicolumn{1}{c|}{SA} & SA \\
		HOME\cite{gilles2021home} & \multicolumn{1}{c|}{CSDTS} & \multicolumn{1}{c|}{BEV} & - & Heatmap, $\uptau_{tar}$ & \multicolumn{1}{c|}{1D-Conv, RNN} & \multicolumn{1}{c|}{CNN} & SA \\
		TPCN\cite{ye2021tpcn} & \multicolumn{1}{c|}{CSDTS} & \multicolumn{1}{c|}{Points} & Tables & $\uptau_{tar}$, $\Delta$ & \multicolumn{2}{c|}{PointNet++, Spatial\&Temporal} & - \\
		LaPred\cite{kim2021lapred} & \multicolumn{1}{c|}{CSDTS} & \multicolumn{1}{c|}{Points} & - & $\uptau_{tar}$ & \multicolumn{2}{c|}{1D-Conv, LSTM, MLP} & SA \\
		MMTrans\cite{liu2021multimodal} & \multicolumn{1}{c|}{CSDTS} & \multicolumn{1}{c|}{$\Vec{Polyline}$} & - & $\uptau_{tar}$ & \multicolumn{1}{c|}{XFMR} & \multicolumn{1}{c|}{VectorNet} & XFMR \\
		ALAN\cite{narayanan2021divide} & \multicolumn{1}{c|}{$\displaystyle CSDTS_{multi}$} & \multicolumn{1}{c|}{Points, BEV} & - & $\uptau_{tar}$ & \multicolumn{1}{c|}{MLP, LSTM} & \multicolumn{1}{c|}{1D-Conv} & S-Pool \\
		LaneRCNN\cite{zeng2021lanercnn} & \multicolumn{1}{c|}{CSDTS} & \multicolumn{1}{c|}{Points, $\mathcal{G}_{Lane}$} & LaneRoI & $\Pr(\cdot)$, $\Delta$, $\Delta\theta$ & \multicolumn{2}{c|}{LaneConv, LanePool} & LaneRoI \\
		PRIME\cite{song2021learning} & \multicolumn{1}{c|}{$\displaystyle CSDTS_{multi}$} & \multicolumn{1}{c|}{$\mathcal{G}_{Lane}$} & $Anchor_T$ & Score & \multicolumn{1}{c|}{1D-Conv, RNN} & \multicolumn{1}{c|}{bi-LSTM} & SA \\
		SceneTrans\cite{ngiam2021scene} & \multicolumn{1}{c|}{CSDTS} & \multicolumn{1}{c|}{$\Vec{Polyline}$} & - & $\uptau_{tar}$, $\mathcal{L}$, $\theta$ & \multicolumn{2}{c|}{XFMR, PointNet\cite{qi2017pointnet}} & SA \\ 
	\bottomrule
    \end{tabular}
\vspace{-6mm}
\end{table*}

\subsection{Representations}
\label{sec:method_repre}

There are mainly two types of representation, i.e., image and continuous-space samples, to describe the historical observation and future prediction in each vehicle trajectory prediction case. 
Images are commonly used to carry the agents and road observations \cite{lee2017desire,cui2019multimodal, chai2019multipath, phan2020covernet, fang2020tpnet,park2020diverse, sriram2020smart, gilles2021home, narayanan2021divide} due to its dense characteristic.
Some researchers \cite{gao2020vectornet, zhao2020tnt, ye2021tpcn, kim2021lapred} prefer utilizing sparse points or polylines when describing the historical trajectories or the scene context.

\subsubsection{Agent State Representations}
Due to the limitation of practical detection and tracking system, the agent state can only be recorded periodically.
It's natural to represent the agent state with continuous-space discrete-time samples (CSDTS), which are vectors (or arrays) containing the agent state features.
In \cite{lee2017desire, gao2020vectornet, mercat2020multi, luo2020probabilistic, pan2020lane, park2020diverse, song2020pip, zhao2020tnt, khandelwal2020if, gilles2021home, ye2021tpcn, kim2021lapred, liu2021multimodal, narayanan2021divide, zeng2021lanercnn, song2021learning, ngiam2021scene}, features mainly include the timestamp, the position of the agent under a bird-eye-view (BEV) coordinate system, the displacement to the last timestamp, the velocity of the agent, and the relative heading angle.
Nevertheless, researchers also explore sketching the trajectories on a rasterized BEV image in \cite{cui2019multimodal, chai2019multipath, phan2020covernet, fang2020tpnet, sriram2020smart}.
The image representation is not widely adopted to describe agent states in recent papers. 

\subsubsection{Scene Context Representations}
In contrast with the continuous-space sample, image is the most straightforward representation to carry the scene context and has been studied in many works \cite{lee2017desire, cui2019multimodal, chai2019multipath, phan2020covernet, fang2020tpnet, park2020diverse, sriram2020smart, gilles2021home, narayanan2021divide}. 
The shape and status of the roads are visualized on the BEV image as detailedly as possible. 
Sometimes, this kind of rasterized image is also called the high definition (HD) map in the paper \cite{gao2020vectornet, zhao2020tnt}.
Gao et al. \cite{gao2020vectornet} propose to discretize the scene context and represent it by the vectors.
This novel idea is quickly appreciated and adopted by other researchers \cite{zhao2020tnt, khandelwal2020if, liu2021multimodal, ngiam2021scene} because it provides a unified representation and the computation cost of vectorized representation is less expensive.
In advance, some works\cite{liang2020learning, ye2021tpcn, zeng2021lanercnn} construct extra tables or graphs to indicate the temporal and spatial correspondence of the vectorized scene context, which assist the feature extraction and interaction modeling in their prediction model. 

\subsubsection{Output Representations}
Some approaches focus on improving the input data representation\cite{gao2020vectornet}, the modeling \cite{luo2020probabilistic, song2020pip, kim2021lapred, liu2021multimodal} as well as the objective function \cite{narayanan2021divide}, and merely output future trajectory represented by single modality, which denotes sets of future position $\uptau_{tar}$ .
In order to produce a promising result, approaches estimate additional modalities, such as the offset of the predicted trajectories from the ground truth \cite{lee2017desire, fang2020tpnet, zeng2021lanercnn}, the score of the prediction according to a pre-defined scoring metric \cite{lee2017desire, liang2020learning, zhao2020tnt, song2021learning}, the probability $\Pr(\cdot)$ of each trajectory prediction or anchor \cite{cui2019multimodal, chai2019multipath, phan2020covernet, fang2020tpnet, zeng2021lanercnn}, the probability heat-map of final position\cite{gilles2021home}, and the heading $\theta$ or the angular offset $\Delta\theta$ of the target agent at the destination\cite{zeng2021lanercnn, ngiam2021scene}.
MultiPath\cite{chai2019multipath} models the control uncertainty with the Gaussian mixture model (GMM) and predicts the mean $\mu$ and variance $\sigma$ of the GMM as the model output.
TPNet\cite{fang2020tpnet} and PRIME\cite{song2021learning} decide to express the trajectory anchors or the output trajectories in the form of polynomials $P(t)$ other than the usual point sets $\uptau_{tar}$. 

\subsection{Modeling}
\label{sec:method_modeling}

To achieve better prediction performance, researchers propose novel models with various architectures, such as Multi-Layer Perception (MLP), Convolution Neural Network (CNN), Recurrent Neural Network (RNN), Graph Neural Network (GNN). 
We summarise some common design choices and their differences regarding feature encoding, interaction modeling, and prediction head.
Some approaches that adopt the generative model are also reviewed. 

\subsubsection{Feature Encoding}
Taking the trajectory as sequential data, plenty of papers \cite{lee2017desire, mercat2020multi, luo2020probabilistic, pan2020lane, park2020diverse, song2020pip, khandelwal2020if} propose feature encoder consisting of RNN units, like the gate recurrent unit (GRU)\cite{cho2014learning} and the long short-term memory. 
Some combine MLP or 1D-Conv with RNN unit to extract hidden space features of the trajectory input\cite{gilles2021home, kim2021lapred, narayanan2021divide, song2021learning} or the scene context input\cite{kim2021lapred}.
The great success of transformer in natural language processing attracts some researchers' attention and they apply the transformer module in feature extraction\cite{khandelwal2020if, liu2021multimodal, ngiam2021scene}.
To handle the rasterized input,  some work\cite{lee2017desire, cui2019multimodal, chai2019multipath, phan2020covernet, park2020diverse, sriram2020smart, gilles2021home} directly borrow convolutional feature encoder from object detection \cite{he2016deep, sandler2018mobilenetv2} and image segmentation tasks\cite{ronneberger2015u}.
VectorNet\cite{gao2020vectornet} follows the idea of PointNet\cite{qi2017pointnet} and extracts instant-level features for the vectorized input.
TNT \cite{zhao2020tnt} directly inherits the feature extraction backbone of VectorNet. 
Ye et al. take the discrete input as the point cloud and mimic the point cloud encoding in their work\cite{ye2021tpcn}.
LaneGCN \cite{liang2020learning} and LaneRCNN\cite{zeng2021lanercnn} introduce graph convolution modules at the encoding stage and aggregate the features across the graph constructed based on the road connectivity or temporal sequence. 

\subsubsection{Interaction Modeling}
The interaction among vehicles, pedestrians and road elements is extremely important yet complicated.
To model the agent-to-agent interaction and agent-to-scene interaction at the same time, researchers enhance the social pooling mechanism proposed in \cite{alahi2016social, gupta2018social} by including the scene context feature maps\cite{lee2017desire, sriram2020smart, narayanan2021divide}.
A variety of researchers\cite{gao2020vectornet, mercat2020multi, luo2020probabilistic, liang2020learning, park2020diverse, khandelwal2020if, gilles2021home, kim2021lapred, song2021learning} motivated by the success of the attention mechanism design the interaction modeling module. 
Furthermore, Liu et al.\cite{liu2021multimodal} and Ngiam et al.\cite{ngiam2021scene} build the entire interaction module with the multi-head attention.
The graph modeling and the GNN layers are also frequently involved because the message passing of GNN can fuse the features of different agents (and road elements). 
We believe the feature fusion behavior is essentially the same as the interaction between agents and the road elements.
There exists an exception that TPCN\cite{ye2021tpcn} doesn't explicitly consider modeling the interaction but still achieves a good performance. 

\subsubsection{Prediction Head}
Some researchers characterise the prediction by hidden Markov model and generate the trajectory prediction with RNN \cite{lee2017desire, mercat2020multi, luo2020probabilistic, park2020diverse, song2020pip, khandelwal2020if}. 
Other researchers take the prediction as a regression process and decode the features with MLP in \cite{gao2020vectornet, cui2019multimodal, pan2020lane, liang2020learning, zhao2020tnt, gilles2021home, kim2021lapred, liu2021multimodal, zeng2021lanercnn, song2021learning, ngiam2021scene}. 
Significantly influenced by the concept of anchor proposal\cite{erhan2014scalable, ren2015faster} in the computer vision field, many researchers add a classification branch in their prediction head to predict a probability or a confidence score for each proposed trajectory anchor \cite{chai2019multipath, fang2020tpnet, luo2020probabilistic, liang2020learning, song2020pip, zhao2020tnt, kim2021lapred, liu2021multimodal, ngiam2021scene, zeng2021lanercnn}.
Motivated by the prediction uncertainty of the deep-learning model against the unseen cases, CoverNet\cite{phan2020covernet} and PRIME\cite{song2021learning} further deploy a model-based planner to propose trajectory anchors that satisfy the constraints imposed by the kinematics of the vehicle and the scene context.

\subsubsection{Generative Model}
DESIRE\cite{lee2017desire} and SMART\cite{sriram2020smart} treat the trajectory prediction as the conditional sampling and selection process.
The recognition module firstly projects the trajectory observation and prediction ground truth to the latent space. 
The latent variables $z$ are assumed to satisfy a distribution prior parameterized by the encoding of the recognition module. 
Then, the conditional decoder recovers the future trajectory from the given conditional input and the latent variable $z$.
However, one cannot obtain the likelihood of each trajectory sampled from the generative model.
DESIRE designs the ranking and refinement module to evaluate the trajectories generated from the CVAE module. 

\subsection{Learning and Objective Functions}
\label{sec:method_learning}
Most of the proposed deep-learning models are trained in a supervised manner.
The cross-entropy (CE) loss, smooth-L1 (Huber) loss, negative log-likelihood (NLL) loss and mean-square-error (MSE) loss are all commonly adopted as the objective function during the training process.
Researchers use some training skills and propose new techniques or objective functions so that the model can converge to a better optimal.

MultiPath\cite{chai2019multipath} adopts an unsupervised learning method to find the appropriate trajectory anchors and teaches their model with imitation learning.
VectorNet\cite{gao2020vectornet} designs an auxiliary graph completion loss to encourage the interaction module to capture a better insight. 
TNT\cite{zhao2020tnt} trains the trajectory regressor by a teacher-forcing technique\cite{williams1989learning} based on their uni-modal assumption.

Models which label the ground truth as one certain trajectory while predicting diverse outputs suffer from mode collapse problem \cite{rhinehart2018r2p2, hong2019rules}, resulting in the failure of making the multi-modal predictions. 
To overcome this problem, some researchers deploy a novel objective function or introduce new functions.
MTP\cite{cui2019multimodal} targets at minimizing the multiple-trajectory prediction loss, which can be regarded as a variant of winner-takes-all (WTA) loss proposed in \cite{lee2016stochastic}:
\begin{equation}
    \mathcal{L}^{MTP}= - \sum_{i=1}^{K} I_{i^*} \log{p_{i}} + \alpha \sum_{i=1}^{K} I_{i^*} L(\tau_{gt}, \tau_{i}),
\end{equation}
where $I_{i^*}$ is a binary indicator equal to 1 if the $i^*$ mode is closest to the GT trajectory according to an arbitrary trajectory distance function or 0 otherwise, $p_i$ is the probability of the best mode $i^*$, $L(\cdot, \cdot)$ indicates an arbitrary displacement error function, and $\alpha$ is a coefficient weighting the regression loss. 
Some followers also incorporate the WTA loss in their work\cite{liang2020learning, khandelwal2020if, ngiam2021scene}.
Narayanan et al.\cite{narayanan2021divide} propose a divide-and-conquer (DAC) initialization technique for stabilizing the training with WTA loss. 
The model trained with DAC beats other competitors in their work.

\section{Implementation of TNT Approach}
\label{sec:imp_tnt}
TNT\cite{zhao2020tnt}, as the extension of VectorNet\cite{gao2020vectornet}, draws our interest with its good model interpretability and outstanding performance. 
However, the official code is disclosed to the public due to the intellectual property issue. 
Aiming at facilitating the exploration of using the vector and graph representation in vehicle trajectory prediction, we implement our version of TNT\cite{zhao2020tnt} with our understanding of the TNT\cite{zhao2020tnt} approach and make our implementation publicly available.

Our code is implemented with Pytorch and Pytorch Geometric libraries. 
Strictly following the details in \cite{gao2020vectornet, zhao2020tnt}, we implement the VectorNet feature extraction backbone and TNT prediction heads. 
The three prediction heads, target candidate prediction head, target-conditioned trajectory prediction head, and scoring head are modeled by 2-layer MLPs. 
The target candidate prediction module consists of two 2-layer MLPs, one for predicting the discrete distribution over target locations, the other one for predicting the most likely offset corresponding to each target candidate.
In our design, the target-conditioned trajectory prediction module is implemented by a 2-layer MLP, but it is involved twice at the loss computation of each batch. 
For the first time, the MLP predicts the trajectories given the ground truth final position regarded as the teacher-forcing training. 
Then, the MLP predicts M trajectories, which will be scored by the scoring module, given the M-selected target candidates.

During the implementation of TNT\cite{zhao2020tnt}, we found the descriptions of several design choices and hyper-parameters are vague or missing. 
We propose proper solutions to fill in these gaps and introduce them briefly in the remaining of this section:

\textbf{Data Normalization}: As mentioned in \cite{gao2020vectornet}, we centralize the coordinates at the last observed position of the target agent for each data sequence. Additionally, we apply the heading normalization as described in \cite{mercat2020multi, zeng2021lanercnn, ngiam2021scene}. The trajectories and scene context representations are rotated to ensure that the heading of the target agent at the last observed position is aligned with the x-axis. 

\textbf{Target Candidates}: According to the description in \cite{zhao2020tnt}, TNT deploys an equal-distance target candidate sampling along with the candidate center lines. 
The authors take $1000$ target candidates as an example. 
It's worth noting that the number of target candidates can vary across the sequences since the possible travel distance and area of the target agent in the future can be diverse. 
A unified number of candidates can fail to cover a sufficient area for predicting the future target.
In our code, we implement the equal-distance sampling strategy and sample different numbers of target candidates for each sequence. 

\textbf{Non-Maximum Suppression (NMS)}: Inspired by object detection, TNT\cite{zhao2020tnt} filters M multi-modal predictions concerning the predicted score and their distance to the selected ones.
The M trajectories are sorted by the predicted score firstly then picked greedily if their distances to the selected ones exceed a certain threshold. 
However, the exact distance threshold that determines the near-duplicate trajectories is not given in their paper. 
Here we implement a hard-threshold strategy. 
Specifically, only trajectories that exceed the distance threshold are selected and all-zero trajectories are padded to the remaining slots if no more trajectory is distant enough from all the selected ones. 

Although exactly the same results as \cite{zhao2020tnt} have not yet been achieved, our implementation still demonstrates outstanding performance.
The results achieved so far are shown in Section \ref{sec:exp}.

\section{Experiments}
\label{sec:exp}
This section is organized as follows. Firstly, several frequently used datasets are introduced. 
Secondly, we present existing metrics that the experiments are evaluated with. In addition, we list the commonly adopted preprocessing tricks and data augmentation strategies. Finally, a detailed comparison of recently proposed methods is demonstrated.

\begin{table*}[t]
\centering
\caption{Trajectory Prediction Performance on Argoverse Dataset}
\begin{adjustbox}{width=1\textwidth}
    \label{tab:argo_results}
    \begin{tabular}{@{}lccccccccccccc@{}}
    \toprule
    \multirow{3}{*}{} & \multicolumn{6}{c}{Validation Set} & \multicolumn{7}{c}{Test Set} \\
    \cmidrule(lr){2-7} \cmidrule(lr){8-14}
    { }& \multicolumn{3}{c}{k=1} & \multicolumn{3}{c}{k=6} & \multicolumn{3}{c}{k=1} & \multicolumn{4}{c}{k=6} \\
    \cmidrule(lr){2-4} \cmidrule(lr){5-7} \cmidrule(lr){8-10} \cmidrule(lr){11-14}
    Model & minADE$^{\dagger}$ & minFDE$^{\dagger}$ & MR$^{\ddagger}$ & minADE & minFDE & MR & minADE & minFDE & MR & minADE & minFDE & MR & DAC \\ \midrule
    VectorNet\cite{gao2020vectornet} & 1.66 & 3.67 & - & - & - & - & 1.81 & 4.01 & - & - & - & - & - \\
    TPNet\cite{fang2020tpnet} & 1.75 & 3.88 & - & - & - & - & 2.23 & 4.70 & - & 1.61 & 3.28 & - & 0.96 \\
    Luo\cite{luo2020probabilistic} & 1.60 & 3.64 & - & 1.35 & 2.68 & - & 1.91 & 4.31 & 0.66 & 0.99 & 1.71 & 0.19 & 0.98 \\
    LaneGCN\cite{liang2020learning} & 1.35 & 2.97 & - & 0.71 & 1.08 & - & 1.71 & 3.78 & 0.59 & 0.87 & 1.36 & 0.16 & - \\
    SMART\cite{sriram2020smart} & - & - & - & 1.44 & 2.47 & - & - & - & - & - & - & - & - \\
    TNT\cite{zhao2020tnt} & - & - & - & 0.73 & 1.29 & 0.09 & - & - & - & 0.94 & 1.54 & 0.13 & - \\
    WIMP\cite{khandelwal2020if} & 1.45 & 3.19 & - & 0.75 & 1.14 & 0.12 & 1.82 & 4.03 & - & 0.90 & 1.42 & 0.17 & - \\
    HOME\cite{gilles2021home} & - & 3.02 & 0.51 & - & 1.28 & 0.07 & 1.73 & 3.73 & 0.58 & 0.94 & 1.45 & \textbf{0.10} & - \\
    TPCN\cite{ye2021tpcn} & 1.34 & 2.95 & 0.50 & 0.73 & 1.15 & 0.11 & \textbf{1.66} & \textbf{3.69} & 0.59 & 0.87 & 1.38 & 0.16 & - \\
    LaPred\cite{kim2021lapred} & 1.48 & 3.29 & - & 0.71 & 1.44 & - & - & - & - & - & - & - & - \\
    MMTrans\cite{liu2021multimodal} & - & - & - & 0.71 & 1.15 & 0.11 & - & - & - & 0.84 & 1.34 & 0.15 & - \\
    LaneRCNN\cite{zeng2021lanercnn} & 1.33 & 2.85 & - & 0.77 & 1.19 & 0.08 & 1.69 & \textbf{3.69} & \textbf{0.57} & 0.90 & 1.45 & 0.12 & - \\
    PRIME\cite{song2021learning} & - & - & - & - & - & - & 1.91 & 3.82 & 0.59 & 1.22 & 1.56 & 0.12 & - \\
    SceneTrans\cite{ngiam2021scene} & - & - & - & - & - & - & - & - & - & \textbf{0.80} & \textbf{1.23} & 0.13 & - \\
    \midrule
    Ours & - & - & - & 1.11 & 2.12 & 0.31 & - & - & - & - & - & - & - \\
    \bottomrule
    \multicolumn{14}{l}{$\star$ minADE/ minFDE: in meters} \\
    \multicolumn{14}{l}{$\ddagger$ MR: the threshold for endpoint error is 2m}
    \end{tabular}
\end{adjustbox}
\vspace{-8mm}
\end{table*}

\subsection{Datasets}
\label{sec:exp_dataset}
\textbf{Argoverse} The Argoverse Dataset\cite{Argoverse} is the most frequently used dataset for motion forecasting experiments. It contains over 300K scenarios in Pittsburgh and Miami. Each scenario contains a 2D birds-eye-view centroid of different objects at 10 Hz. The task of motion forecasting is to predict the trajectories of the sole agent type object in the next 3 seconds, given the preceding 2-second long trajectories together with HD map features which can be accessed through Argoverse API. The whole dataset can be split into 208,272 training sequences, 40,127 validation sequences, and 79,391 testing sequences. For training and validation, full 5-second trajectories are provided. While for testing, only the first 2 seconds trajectories are given.

\textbf{NuScenes} NuScenes is a public large-scale dataset for autonomous driving\cite{nuscene}. The trajectories are represented in the x-y coordinate system at 2 Hz. The original driving scenarios are collected in Boston and Singapore, where right-hand and left-hand traffic rules apply respectively. Up to 2 seconds of past history can be utilized to predict 6-second long future trajectories for each agent.

\textbf{NGSIM} Next Generation SIMulation\cite{NGSIM} dataset is extracted from real-world highway driving scenarios using high mounted digital video cameras. The precise location of each vehicle is recorded at 10 Hz. Since the dataset is not published purely for trajectory prediction, the length of past observations and predictions can be customized.

\subsection{Metrics}
\label{sec:exp_metric}

Some commonly used metrics are as follows.

\textbf{FDE(K)} Minimum Final Displacement Error over K refers to the L2 distance between the endpoint of predicted trajectory and that of ground truth over the best K predictions. 

\textbf{ADE(K)} Minimum Average Displacement Error over K denotes the average pointwise L2 distance between the whole forecasted trajectory and the ground truth over the best K predictions. 

\textbf{MR} Miss Rate evaluates the proportion of unacceptable outputs in all proposed solutions. One scenario is usually defined as a miss when the endpoint error for the best trajectory is greater than 2.0m.

Some papers have proposed several specific metrics corresponding to their innovative methods.

\textbf{DAC} Drivable Area Compliance\cite{fang2020tpnet,luo2020probabilistic,park2020diverse} equals to the count of future trajectories within the drivable area divided by the number of all possible trajectories. DAC evaluates the feasibility of proposed solutions.

\subsection{Preprocessing and Augmentation}
\label{sec:exp_pre}

\textbf{Centralization} Aiming at reducing the complexity of prediction, the origin of the coordinate system to represent trajectories is chosen to be the position of the predicted agent at the last observed timestamp\cite{phan2020covernet,gao2020vectornet,mercat2020multi,zeng2021lanercnn,ngiam2021scene}.

\textbf{Heading Normalization} Heading normalization means that the coordinate system where trajectories are represented is rotated\cite{phan2020covernet,mercat2020multi,zeng2021lanercnn,ngiam2021scene}, such that the orientation of the predicted agent at the last observed timestamp is aligned with the x-axis. 

\textbf{Scene Rotation} In order to combat overfitting and improve generalization capability, random rotation\cite{lee2017desire,gilles2021home,zeng2021lanercnn,ngiam2021scene} is frequently applied to the whole scenario and the trajectory coordinates following centralization.

\textbf{Agent Dropout} As is often the case, there are excessive agents in one scenario, some of which are not worthy of attention. In \cite{ngiam2021scene}, non-predicted agents are artificially removed with probability of 0.1.

\subsection{Results}
\label{sec:exp_res}
This section makes a comparison among the results of the fore-mentioned methods. Since Argoverse is the most frequently used data set, for a fair comparison we only compare experimental results in Argoverse. All data is collected originally from the published papers and listed in TABLE \ref{tab:argo_results}.

Scene Transformer\cite{ngiam2021scene} ranks first when sorting results by minADE(K=6) and minFDE(K=6), while the first prize belongs to \cite{gilles2021home} when ranked by MR(K=6). Both methods use the attention module to model the interaction between agents and environments, which provides a promising solution to interaction representation.
A performance drop is usually observed among other approaches (TNT\cite{zhao2020tnt}, WIMP\cite{khandelwal2020if}, HOME\cite{gilles2021home}, TPCN\cite{ye2021tpcn}, MMTrans\cite{liu2021multimodal}, annd laneRCNN\cite{zeng2021lanercnn}) when applying them on the test set.
We evaluate our TNT implementation on the validation set of Argoverse. 
Compared with the state-of-the-art, our performance is 0.4 meter worse in minADE, 1.04 meters in minFDE, 0.24 in MR.
Improvement is expected by introducing the WTA loss or the lane attention strategy in our design. 

\section{Conclusion}
\label{sec:conclusion}
In this work, we make a thorough review of existing methods on trajectory prediction for vehicles. Starting from the mathematical formulation of the prediction problem, we divide the complicated forecasting task into three components: representation, modeling as well as learning and objective functions. Modules varying from MLP to CNN, RNN, GNN form constituent parts of prediction networks, namely feature encoding, interaction modeling, and prediction header. We additionally make our version of TNT publicly available and present the implementation in detail. A fair comparison of results on Argoverse Dataset alongside with list of metrics and preprocessing tricks are illustrated at the end of this work. Further improvement of trajectory prediction, e.g., accuracy or efficiency, is expected with the assistance of our work.

\section*{Acknowledgment}

This work is partially supported by Shenzhen Key Laboratory of Robotics Perception and Intelligence, Southern University of Science and Technology, Shenzhen 518055, China, Hong Kong RGC CRF grant C4063-18G, Hong Kong RGC GRF grant \#14200618.

%
%

\bibliographystyle{IEEEtran}
\bibliography{reference_jb} 

\begin{thebibliography}{10}
\providecommand{\url}[1]{#1}
\csname url@samestyle\endcsname
\providecommand{\newblock}{\relax}
\providecommand{\bibinfo}[2]{#2}
\providecommand{\BIBentrySTDinterwordspacing}{\spaceskip=0pt\relax}
\providecommand{\BIBentryALTinterwordstretchfactor}{4}
\providecommand{\BIBentryALTinterwordspacing}{\spaceskip=\fontdimen2\font plus
\BIBentryALTinterwordstretchfactor\fontdimen3\font minus
  \fontdimen4\font\relax}
\providecommand{\BIBforeignlanguage}[2]{{%
\expandafter\ifx\csname l@#1\endcsname\relax
\typeout{** WARNING: IEEEtran.bst: No hyphenation pattern has been}%
\typeout{** loaded for the language `#1'. Using the pattern for}%
\typeout{** the default language instead.}%
\else
\language=\csname l@#1\endcsname
\fi
#2}}
\providecommand{\BIBdecl}{\relax}
\BIBdecl

\bibitem{ess2010object}
A.~Ess, K.~Schindler, B.~Leibe, and L.~Van~Gool, ``Object detection and
  tracking for autonomous navigation in dynamic environments,'' \emph{The
  International Journal of Robotics Research}, vol.~29, no.~14, pp. 1707--1725,
  2010.

\bibitem{pellegrini2011predicting}
S.~Pellegrini, A.~Ess, and L.~Van~Gool, ``Predicting pedestrian trajectories,''
  in \emph{Visual Analysis of Humans}.\hskip 1em plus 0.5em minus 0.4em\relax
  Springer, 2011, pp. 473--491.

\bibitem{luber2010people}
M.~Luber, J.~A. Stork, G.~D. Tipaldi, and K.~O. Arras, ``People tracking with
  human motion predictions from social forces,'' in \emph{2010 IEEE
  International Conference on Robotics and Automation}.\hskip 1em plus 0.5em
  minus 0.4em\relax IEEE, 2010, pp. 464--469.

\bibitem{zhao2020tnt}
H.~Zhao, J.~Gao, T.~Lan, C.~Sun, B.~Sapp, B.~Varadarajan, Y.~Shen, Y.~Shen,
  Y.~Chai, C.~Schmid \emph{et~al.}, ``Tnt: Target-driven trajectory
  prediction,'' \emph{arXiv preprint arXiv:2008.08294}, 2020.

\bibitem{alahi2016social}
A.~Alahi, K.~Goel, V.~Ramanathan, A.~Robicquet, L.~Fei-Fei, and S.~Savarese,
  ``Social lstm: Human trajectory prediction in crowded spaces,'' in
  \emph{Proceedings of the IEEE conference on computer vision and pattern
  recognition}, 2016, pp. 961--971.

\bibitem{gupta2018social}
A.~Gupta, J.~Johnson, L.~Fei-Fei, S.~Savarese, and A.~Alahi, ``Social gan:
  Socially acceptable trajectories with generative adversarial networks,'' in
  \emph{Proceedings of the IEEE Conference on Computer Vision and Pattern
  Recognition}, 2018, pp. 2255--2264.

\bibitem{ye2021tpcn}
M.~Ye, T.~Cao, and Q.~Chen, ``Tpcn: Temporal point cloud networks for motion
  forecasting,'' in \emph{Proceedings of the IEEE/CVF Conference on Computer
  Vision and Pattern Recognition}, 2021, pp. 11\,318--11\,327.

\bibitem{lee2017desire}
N.~Lee, W.~Choi, P.~Vernaza, C.~B. Choy, P.~H. Torr, and M.~Chandraker,
  ``Desire: Distant future prediction in dynamic scenes with interacting
  agents,'' in \emph{Proceedings of the IEEE Conference on Computer Vision and
  Pattern Recognition}, 2017, pp. 336--345.

\bibitem{cui2019multimodal}
H.~Cui, V.~Radosavljevic, F.-C. Chou, T.-H. Lin, T.~Nguyen, T.-K. Huang,
  J.~Schneider, and N.~Djuric, ``Multimodal trajectory predictions for
  autonomous driving using deep convolutional networks,'' in \emph{2019
  International Conference on Robotics and Automation (ICRA)}.\hskip 1em plus
  0.5em minus 0.4em\relax IEEE, 2019, pp. 2090--2096.

\bibitem{sandler2018mobilenetv2}
M.~Sandler, A.~Howard, M.~Zhu, A.~Zhmoginov, and L.-C. Chen, ``Mobilenetv2:
  Inverted residuals and linear bottlenecks,'' in \emph{Proceedings of the IEEE
  conference on computer vision and pattern recognition}, 2018, pp. 4510--4520.

\bibitem{chai2019multipath}
Y.~Chai, B.~Sapp, M.~Bansal, and D.~Anguelov, ``Multipath: Multiple
  probabilistic anchor trajectory hypotheses for behavior prediction,''
  \emph{arXiv preprint arXiv:1910.05449}, 2019.

\bibitem{he2016deep}
K.~He, X.~Zhang, S.~Ren, and J.~Sun, ``Deep residual learning for image
  recognition,'' in \emph{Proceedings of the IEEE conference on computer vision
  and pattern recognition}, 2016, pp. 770--778.

\bibitem{phan2020covernet}
T.~Phan-Minh, E.~C. Grigore, F.~A. Boulton, O.~Beijbom, and E.~M. Wolff,
  ``Covernet: Multimodal behavior prediction using trajectory sets,'' in
  \emph{Proceedings of the IEEE/CVF Conference on Computer Vision and Pattern
  Recognition}, 2020, pp. 14\,074--14\,083.

\bibitem{gao2020vectornet}
J.~Gao, C.~Sun, H.~Zhao, Y.~Shen, D.~Anguelov, C.~Li, and C.~Schmid,
  ``Vectornet: Encoding hd maps and agent dynamics from vectorized
  representation,'' in \emph{Proceedings of the IEEE/CVF Conference on Computer
  Vision and Pattern Recognition}, 2020, pp. 11\,525--11\,533.

\bibitem{qi2017pointnet}
C.~R. Qi, H.~Su, K.~Mo, and L.~J. Guibas, ``Pointnet: Deep learning on point
  sets for 3d classification and segmentation,'' in \emph{Proceedings of the
  IEEE conference on computer vision and pattern recognition}, 2017, pp.
  652--660.

\bibitem{fang2020tpnet}
L.~Fang, Q.~Jiang, J.~Shi, and B.~Zhou, ``Tpnet: Trajectory proposal network
  for motion prediction,'' in \emph{Proceedings of the IEEE/CVF Conference on
  Computer Vision and Pattern Recognition}, 2020, pp. 6797--6806.

\bibitem{mercat2020multi}
J.~Mercat, T.~Gilles, N.~El~Zoghby, G.~Sandou, D.~Beauvois, and G.~P. Gil,
  ``Multi-head attention for multi-modal joint vehicle motion forecasting,'' in
  \emph{2020 IEEE International Conference on Robotics and Automation
  (ICRA)}.\hskip 1em plus 0.5em minus 0.4em\relax IEEE, 2020, pp. 9638--9644.

\bibitem{luo2020probabilistic}
C.~Luo, L.~Sun, D.~Dabiri, and A.~Yuille, ``Probabilistic multi-modal
  trajectory prediction with lane attention for autonomous vehicles,'' in
  \emph{2020 IEEE/RSJ International Conference on Intelligent Robots and
  Systems (IROS)}.\hskip 1em plus 0.5em minus 0.4em\relax IEEE, 2020, pp.
  2370--2376.

\bibitem{pan2020lane}
J.~Pan, H.~Sun, K.~Xu, Y.~Jiang, X.~Xiao, J.~Hu, and J.~Miao, ``Lane-attention:
  Predicting vehicles’ moving trajectories by learning their attention over
  lanes,'' in \emph{2020 IEEE/RSJ International Conference on Intelligent
  Robots and Systems (IROS)}.\hskip 1em plus 0.5em minus 0.4em\relax IEEE,
  2020, pp. 7949--7956.

\bibitem{liang2020learning}
M.~Liang, B.~Yang, R.~Hu, Y.~Chen, R.~Liao, S.~Feng, and R.~Urtasun, ``Learning
  lane graph representations for motion forecasting,'' in \emph{European
  Conference on Computer Vision}.\hskip 1em plus 0.5em minus 0.4em\relax
  Springer, 2020, pp. 541--556.

\bibitem{park2020diverse}
S.~H. Park, G.~Lee, J.~Seo, M.~Bhat, M.~Kang, J.~Francis, A.~Jadhav, P.~P.
  Liang, and L.-P. Morency, ``Diverse and admissible trajectory forecasting
  through multimodal context understanding,'' in \emph{European Conference on
  Computer Vision}.\hskip 1em plus 0.5em minus 0.4em\relax Springer, 2020, pp.
  282--298.

\bibitem{sriram2020smart}
N.~Sriram, B.~Liu, F.~Pittaluga, and M.~Chandraker, ``Smart: Simultaneous
  multi-agent recurrent trajectory prediction,'' in \emph{European Conference
  on Computer Vision}.\hskip 1em plus 0.5em minus 0.4em\relax Springer, 2020,
  pp. 463--479.

\bibitem{song2020pip}
H.~Song, W.~Ding, Y.~Chen, S.~Shen, M.~Y. Wang, and Q.~Chen, ``Pip:
  Planning-informed trajectory prediction for autonomous driving,'' in
  \emph{European Conference on Computer Vision}.\hskip 1em plus 0.5em minus
  0.4em\relax Springer, 2020, pp. 598--614.

\bibitem{khandelwal2020if}
S.~Khandelwal, W.~Qi, J.~Singh, A.~Hartnett, and D.~Ramanan, ``What-if motion
  prediction for autonomous driving,'' \emph{arXiv preprint arXiv:2008.10587},
  2020.

\bibitem{gilles2021home}
T.~Gilles, S.~Sabatini, D.~Tsishkou, B.~Stanciulescu, and F.~Moutarde, ``Home:
  Heatmap output for future motion estimation,'' \emph{arXiv preprint
  arXiv:2105.10968}, 2021.

\bibitem{kim2021lapred}
B.~Kim, S.~H. Park, S.~Lee, E.~Khoshimjonov, D.~Kum, J.~Kim, J.~S. Kim, and
  J.~W. Choi, ``Lapred: Lane-aware prediction of multi-modal future
  trajectories of dynamic agents,'' in \emph{Proceedings of the IEEE/CVF
  Conference on Computer Vision and Pattern Recognition}, 2021, pp.
  14\,636--14\,645.

\bibitem{liu2021multimodal}
Y.~Liu, J.~Zhang, L.~Fang, Q.~Jiang, and B.~Zhou, ``Multimodal motion
  prediction with stacked transformers,'' in \emph{Proceedings of the IEEE/CVF
  Conference on Computer Vision and Pattern Recognition}, 2021, pp. 7577--7586.

\bibitem{narayanan2021divide}
S.~Narayanan, R.~Moslemi, F.~Pittaluga, B.~Liu, and M.~Chandraker,
  ``Divide-and-conquer for lane-aware diverse trajectory prediction,'' in
  \emph{Proceedings of the IEEE/CVF Conference on Computer Vision and Pattern
  Recognition}, 2021, pp. 15\,799--15\,808.

\bibitem{zeng2021lanercnn}
W.~Zeng, M.~Liang, R.~Liao, and R.~Urtasun, ``Lanercnn: Distributed
  representations for graph-centric motion forecasting,'' \emph{arXiv preprint
  arXiv:2101.06653}, 2021.

\bibitem{song2021learning}
H.~Song, D.~Luan, W.~Ding, M.~Y. Wang, and Q.~Chen, ``Learning to predict
  vehicle trajectories with model-based planning,'' \emph{arXiv preprint
  arXiv:2103.04027}, 2021.

\bibitem{ngiam2021scene}
J.~Ngiam, B.~Caine, V.~Vasudevan, Z.~Zhang, H.-T.~L. Chiang, J.~Ling,
  R.~Roelofs, A.~Bewley, C.~Liu, A.~Venugopal \emph{et~al.}, ``Scene
  transformer: A unified multi-task model for behavior prediction and
  planning,'' \emph{arXiv preprint arXiv:2106.08417}, 2021.

\bibitem{cho2014learning}
K.~Cho, B.~Van~Merri{\"e}nboer, C.~Gulcehre, D.~Bahdanau, F.~Bougares,
  H.~Schwenk, and Y.~Bengio, ``Learning phrase representations using rnn
  encoder-decoder for statistical machine translation,'' \emph{arXiv preprint
  arXiv:1406.1078}, 2014.

\bibitem{ronneberger2015u}
O.~Ronneberger, P.~Fischer, and T.~Brox, ``U-net: Convolutional networks for
  biomedical image segmentation,'' in \emph{International Conference on Medical
  image computing and computer-assisted intervention}.\hskip 1em plus 0.5em
  minus 0.4em\relax Springer, 2015, pp. 234--241.

\bibitem{erhan2014scalable}
D.~Erhan, C.~Szegedy, A.~Toshev, and D.~Anguelov, ``Scalable object detection
  using deep neural networks,'' in \emph{Proceedings of the IEEE conference on
  computer vision and pattern recognition}, 2014, pp. 2147--2154.

\bibitem{ren2015faster}
S.~Ren, K.~He, R.~Girshick, and J.~Sun, ``Faster r-cnn: Towards real-time
  object detection with region proposal networks,'' \emph{Advances in neural
  information processing systems}, vol.~28, pp. 91--99, 2015.

\bibitem{williams1989learning}
R.~J. Williams and D.~Zipser, ``A learning algorithm for continually running
  fully recurrent neural networks,'' \emph{Neural computation}, vol.~1, no.~2,
  pp. 270--280, 1989.

\bibitem{rhinehart2018r2p2}
N.~Rhinehart, K.~M. Kitani, and P.~Vernaza, ``R2p2: A reparameterized
  pushforward policy for diverse, precise generative path forecasting,'' in
  \emph{Proceedings of the European Conference on Computer Vision (ECCV)},
  2018, pp. 772--788.

\bibitem{hong2019rules}
J.~Hong, B.~Sapp, and J.~Philbin, ``Rules of the road: Predicting driving
  behavior with a convolutional model of semantic interactions,'' in
  \emph{Proceedings of the IEEE/CVF Conference on Computer Vision and Pattern
  Recognition}, 2019, pp. 8454--8462.

\bibitem{lee2016stochastic}
S.~Lee, S.~P.~S. Prakash, M.~Cogswell, V.~Ranjan, D.~Crandall, and D.~Batra,
  ``Stochastic multiple choice learning for training diverse deep ensembles,''
  in \emph{Advances in Neural Information Processing Systems}, 2016, pp.
  2119--2127.

\bibitem{Argoverse}
M.-F. Chang, J.~W. Lambert, P.~Sangkloy, J.~Singh, S.~Bak, A.~Hartnett,
  D.~Wang, P.~Carr, S.~Lucey, D.~Ramanan, and J.~Hays, ``Argoverse: 3d tracking
  and forecasting with rich maps,'' in \emph{Conference on Computer Vision and
  Pattern Recognition (CVPR)}, 2019.

\bibitem{nuscene}
H.~Caesar, V.~Bankiti, A.~H. Lang, S.~Vora, V.~E. Liong, Q.~Xu, A.~Krishnan,
  Y.~Pan, G.~Baldan, and O.~Beijbom, ``nuscenes: A multimodal dataset for
  autonomous driving,'' in \emph{Proceedings of the IEEE/CVF conference on
  computer vision and pattern recognition}, 2020, pp. 11\,621--11\,631.

\bibitem{NGSIM}
J.~C. John~Halkias, ``Next generation simulation fact sheet,'' Federal Highway
  Administration (FHWA), 2006.

\end{thebibliography}

\end{document}